\documentclass{article} 

\newcommand{\eg}{{\it e.g.}}
\newcommand{\ie}{{\it i.e.}}

\newcommand{\BA}{\begin{array}}
\newcommand{\EA}{\end{array}}

\newcommand{\BIT}{\begin{itemize}}
\newcommand{\EIT}{\end{itemize}}
\newcommand{\ones}{\mathbf 1}

\newcommand{\reals}{{\mathbb{R}}} 

\newcommand{\Expect}{\mathop{\bf E{}}}

\usepackage{iclr2017_conference,times}
\usepackage{url}            
\usepackage{amsfonts}       
\usepackage{subcaption}

\usepackage{graphicx}

\newcommand{\T}{\mathcal{T}}
\newcommand{\X}{\mathcal{S}}
\newcommand{\A}{\mathcal{A}}

\iclrfinalcopy 
\begin{document}

\title{Combining policy gradient and Q-learning}
\author{Brendan O'Donoghue, R\'emi Munos, Koray Kavukcuoglu \& Volodymyr Mnih
\\ Deepmind \\
\texttt{\{bodonoghue,munos,korayk,vmnih\}@google.com}
}

\maketitle

\begin{abstract}
Policy gradient is an efficient technique for improving a policy in a
reinforcement learning setting. However, vanilla online variants are on-policy
only and not able to take advantage of off-policy data.  In this paper we
describe a new technique that combines policy gradient with off-policy
Q-learning, drawing experience from a replay buffer. This is motivated by making
a connection between the fixed points of the regularized policy gradient
algorithm and the Q-values. This connection allows us to estimate the Q-values
from the action preferences of the policy, to which we apply Q-learning updates.
We refer to the new technique as `PGQL', for policy gradient and
Q-learning. We also establish an equivalency between
action-value fitting techniques and actor-critic algorithms, showing that
regularized policy gradient techniques can be interpreted as advantage function
learning algorithms.  We conclude with some numerical examples that demonstrate
improved data efficiency and stability of PGQL. In particular, we tested
PGQL on the full suite of Atari games and achieved performance exceeding that
of both asynchronous advantage actor-critic (A3C) and Q-learning.
\end{abstract}


\section{Introduction}

In reinforcement learning an agent explores an environment and through the use
of a reward signal learns to optimize its behavior to maximize the expected
long-term return. Reinforcement learning has seen success in several areas
including robotics \citep{lin1993, levine2015endtoend}, computer games
\citep{mnih-atari-2013, mnih-dqn-2015}, online advertising
\citep{pednault2002sequential}, board games \citep{tesauro1995temporal,
silver2016mastering}, and many others.  For an introduction to reinforcement
learning we refer to the classic text by \citet{sutton:book}.  In this paper we
consider model-free reinforcement learning, where the state-transition function
is not known or learned.  There are many different algorithms for model-free
reinforcement learning, but most fall into one of two families:
action-value fitting and policy gradient techniques.

Action-value techniques involve fitting a function, called the Q-values, that
captures the expected return for taking a particular action at a particular
state, and then following a particular policy
thereafter. Two alternatives we discuss in this paper are SARSA
\citep{rummery1994sarsa} and Q-learning \citep{watkins1989learning}, although
there are many others. SARSA is an on-policy algorithm whereby the action-value
function is fit to the current policy, which is then refined by being mostly
greedy with respect to those action-values. On the other hand, Q-learning
attempts to find the Q-values associated with the optimal policy directly and
does not fit to the policy that was used to generate the data.
Q-learning is an off-policy algorithm that can use data generated by another
agent or from a replay buffer of old experience. Under certain conditions both
SARSA and Q-learning can be shown to converge to the optimal Q-values, from
which we can derive the optimal policy \citep{sutton1988learning,
bertsekas1996neuro}.

In policy gradient techniques the policy is represented explicitly and we
improve the policy by updating the parameters in the direction of the gradient
of the performance \citep{sutton1999policy, silver2014deterministic,
kakade2001natural}.  Online policy gradient typically requires an estimate of the
action-value function of the current policy. For this reason they are often
referred to as actor-critic methods, where the actor refers to the policy and
the critic to the estimate of the action-value function
\citep{konda2003onactor}.  Vanilla actor-critic methods are on-policy only,
although some attempts have been made to extend them to off-policy data
\citep{degris2012off, levine2013guided}.

In this paper we derive a link between the Q-values induced by a policy and the
policy itself when the policy is the fixed point of a regularized policy
gradient algorithm (where the gradient vanishes).  This connection allows us to
derive an estimate of the Q-values from the current policy, which we can refine
using off-policy data and Q-learning. We show in the tabular setting that when
the regularization penalty is small (the usual case) the resulting policy is
close to the policy that would be found without the addition of the Q-learning
update. Separately, we show that regularized actor-critic methods can be
interpreted as action-value fitting methods, where the Q-values have been
parameterized in a particular way.  We conclude with some numerical examples
that provide empirical evidence of improved data efficiency and stability of
PGQL.

\subsection{Prior work}
Here we highlight various axes along which our work can be compared to others.
In this paper we use entropy regularization to ensure exploration in the policy,
which is a common practice in policy gradient \citep{williams1991function,
mnih2016asynchronous}. An alternative is to use KL-divergence instead of entropy
as a regularizer, or as a constraint on how much deviation is permitted from a
prior policy \citep{bagnell2003covariant, peters2010relative, schulman2015trust,
glearning}.  Natural policy gradient can also be interpreted as putting a
constraint on the KL-divergence at each step of the policy improvement
\citep{amari1998natural, kakade2001natural, pascanu2013revisiting}.  In
\citet{sallans2004reinforcement} the authors use a Boltzmann exploration policy
over estimated Q-values which they update using TD-learning. In
\citet{heess2012actor} this was extended to use an actor-critic algorithm
instead of TD-learning, however the two updates were not combined as we have
done in this paper.  In \citet{azar2012dynamic} the authors develop an algorithm
called dynamic policy programming, whereby they apply a Bellman-like update to
the action-preferences of a policy, which is similar in spirit to the update we
describe here.  In \citet{norouzi2016reward} the authors augment a maximum
likelihood objective with a reward in a supervised learning setting, and develop
a connection that resembles the one we develop here between the policy and the
Q-values. Other works have attempted to combine on and off-policy learning,
primarily using action-value fitting methods \citep{wang2013backward,
hausknecht2016on, lehnert2015}, with varying degrees of success.  In this paper
we establish a connection between actor-critic algorithms and action-value
learning algorithms.  In particular we show that TD-actor-critic
\citep{konda2003onactor} is equivalent to expected-SARSA \cite[Exercise
6.10]{sutton:book} with Boltzmann exploration where the Q-values are decomposed
into advantage function and value function. The algorithm we develop extends
actor-critic with a Q-learning style update that, due to the decomposition of
the Q-values, resembles the update of the dueling architecture
\citep{wang2015dueling}.  Recently, the field of deep reinforcement learning,
\ie, the use of deep neural networks to represent action-values or a policy, has
seen a lot of success \citep{mnih-dqn-2015, mnih2016asynchronous,
silver2016mastering, riedmiller2005nfq, lillicrap2015continuous,
hado2015doubledqn}. In the examples section we use a neural network with PGQL to
play the Atari games suite.

\section{Reinforcement Learning}
We consider the infinite horizon, discounted, finite state and action space
Markov decision process, with state space $\X$, action space $\A$ and rewards at
each time period denoted by $r_t \in \reals$.  A \emph{policy} $\pi : \X \times
\A \rightarrow \reals_+$ is a mapping from state-action pair to the probability
of taking that action at that state, so it must satisfy $\sum_{a \in \A} \pi(s,
a) = 1$ for all states $s \in \X$. Any
policy $\pi$ induces a probability distribution over visited states, $d^\pi : \X
\rightarrow \reals_+$ (which may depend on the initial state), so the probability
of seeing state-action pair $(s,a) \in \X \times \A$ is $d^\pi(s) \pi(s, a)$.

In reinforcement learning an `agent' interacts with an environment over a number
of times steps. At each time step $t$ the agent receives a state $s_t$ and a
reward $r_t$ and selects an action $a_t$ from the policy $\pi_t$, at which point
the agent moves to the next state
$s_{t+1} \sim P(\cdot, s_t, a_t)$, where $P(s^\prime, s, a)$ is the probability
of transitioning from state $s$ to state $s^\prime$ after taking action $a$.
This continues until the agent encounters a terminal state (after which the
process is typically restarted). The goal of the
agent is to find a policy $\pi$ that maximizes the expected total discounted
return $J(\pi) = \Expect (\sum_{t=0}^\infty \gamma^t r_t \mid \pi)$, where the
expectation is with respect to the initial state distribution, the
state-transition probabilities, and the policy, and where $\gamma \in (0,1)$ is
the discount factor
that, loosely speaking, controls how much the agent prioritizes long-term versus
short-term rewards.  Since the agent starts with no knowledge of the environment
it must continually explore the state space and so will typically use a
stochastic policy.



\paragraph{Action-values.}
The action-value, or Q-value, of a particular state under policy $\pi$ is the
expected total discounted return from taking that action at that state and
following $\pi$ thereafter, \ie, $Q^\pi(s,a) = \Expect( \sum_{t=0}^\infty
\gamma^t r_t \mid s_0 = s, a_0 =
a, \pi)$. The value of state $s$ under policy $\pi$ is denoted by $V^\pi(s) =
\Expect(\sum_{t=0}^\infty \gamma^t r_t \mid s_0 = s, \pi)$, which is the
expected total discounted return of policy $\pi$ from state $s$. The optimal
action-value function
is denoted $Q^\star$ and satisfies $Q^\star(s,a) = \max_\pi Q^\pi(s,a)$
for each $(s,a)$. The policy that achieves the maximum is the optimal policy
$\pi^\star$, with value function $V^\star$.  The advantage function is the
difference between the action-value and the value function, \ie, $A^\pi(s,a) =
Q^\pi(s,a) - V^\pi(s)$, and represents the additional expected reward of taking
action $a$ over the average performance of the policy from state $s$. Since
$V^\pi(s) = \sum_a \pi(s,a) Q^\pi(s,a)$ we have the identity $ \sum_a \pi(s, a)
A^\pi(s,a) = 0$, which simply states that the policy $\pi$ has no advantage
over itself.

\paragraph{Bellman equation.}
The Bellman operator $\T^\pi$ \citep{bellman} for policy $\pi$ is defined as
\[
  \T^\pi Q(s, a) = \Expect_{s^\prime, r, b}(r(s,a) + \gamma
Q(s^\prime, b)),
\]
where the expectation is over next state $s^\prime \sim P(\cdot, s, a)$, the
reward $r(s,a)$, and the
action $b$ from policy $\pi_{s^\prime}$.
The Q-value function for policy $\pi$ is the fixed point
of the Bellman operator
for $\pi$, \ie, $\T^\pi Q^\pi = Q^\pi$. The optimal Bellman operator $\T^\star$
is defined as
\[
\T^\star Q(s, a) = \Expect_{s^\prime, r}(r(s,a) + \gamma \max_b
Q(s^\prime, b)),
\]
where the expectation is over the next state $s^\prime \sim P(\cdot, s, a)$, and
the reward $r(s,a)$.  The optimal Q-value function is the fixed point of the
optimal Bellman equation, \ie,
$\T^\star Q^\star = Q^\star$.
Both the $\pi$-Bellman operator and the optimal Bellman operator are
$\gamma$-contraction mappings in the sup-norm, \ie,
$\|\T Q_1 - \T Q_2\|_\infty \leq \gamma \|Q_1 - Q_2\|_\infty$,
for any $Q_1, Q_2 \in \reals^{\X \times \A}$.
From this fact one can show that the fixed point of each operator is
unique, and that value iteration converges, \ie, $(\T^\pi)^k Q \rightarrow
Q^\pi$ and $(\T^\star)^k Q \rightarrow Q^\star$ from any initial $Q$.
\citep{bertsekas2005dynamic}.

\subsection{Action-value learning}

In value based reinforcement learning we approximate the Q-values using a function
approximator. We then update the parameters so that the Q-values are as close to
the fixed point of a Bellman equation as possible. If we denote by $Q(s, a;\theta)$
the approximate Q-values parameterized by $\theta$, then Q-learning updates the
Q-values along direction $\Expect_{s,a}(\T^\star Q(s,a;\theta) -
Q(s,a;\theta))\nabla_\theta Q(s,a;\theta)$ and SARSA updates the
Q-values along direction $\Expect_{s,a}(\T^\pi Q(s,a;\theta) - Q(s,a;\theta))
\nabla_\theta Q(s,a;\theta)$.  In the online
setting the Bellman operator is approximated by sampling and bootstrapping,
whereby the Q-values at any state are updated using the Q-values from the next
visited state.  Exploration is achieved by not always taking the action with the
highest Q-value at each time step. One common technique called `epsilon greedy' is
to sample a random action with probability $\epsilon > 0$, where $\epsilon$
starts high and decreases over time. Another popular technique is `Boltzmann
exploration', where the policy is given by the softmax over the Q-values with a
temperature $T$, \ie, $\pi(s,a) = \exp(Q(s,a) /T) / \sum_b \exp(Q(s,b) / T)$,
where it is common to decrease the temperature over time.

\subsection{Policy gradient}
\label{s-pg}
Alternatively, we can parameterize the policy directly and attempt to improve it
via gradient ascent on the performance $J$. The policy gradient theorem
\citep{sutton1999policy} states that the gradient of $J$ with respect to the
parameters of the policy is given by
\begin{equation}
  \label{e-pol-grad}
  \nabla_\theta J(\pi) = \Expect_{s,a} Q^\pi(s,a) \nabla_\theta \log \pi(s,a),
\end{equation}
where the expectation is over $(s,a)$ with probability $d^\pi(s)\pi(s,a)$.
In the original derivation of the policy gradient theorem the expectation is
over the \emph{discounted} distribution of states, \ie, over $d_\gamma^{\pi,
s_0}(s) = \sum_{t=0}^\infty \gamma^t Pr \{s_t = s \mid s_0, \pi\}$. However, the
gradient update in that case will assign a low weight to states that take a long
time to reach and can therefore have poor empirical performance.  In practice
the non-discounted distribution of states is frequently used instead.  In
certain cases this is equivalent to maximizing the average (\ie, non-discounted)
policy performance, even when $Q^\pi$ uses a discount factor
\citep{Thomas2014bias}.  Throughout this paper we will use the
\emph{non-discounted} distribution of states.

In the online case it is common to add an entropy regularizer to the gradient in
order to prevent the policy becoming deterministic.  This ensures that the agent
will explore continually. In that case the (batch) update becomes
\begin{equation}
\label{e-rel-pol-grad}
\Delta \theta \propto \Expect_{s,a} Q^\pi(s, a) \nabla_\theta \log\pi(s,a) +
\alpha \Expect_s \nabla_\theta H^\pi(s),
\end{equation}
where $H^\pi(s) = -\sum_a \pi(s, a) \log \pi(s, a)$ denotes the entropy of
policy $\pi$, and $\alpha > 0$ is the
regularization penalty parameter.
Throughout this paper we will make use of entropy regularization, however many
of the results are true for other choices of regularizers with only minor
modification, \eg, KL-divergence.
Note that equation (\ref{e-rel-pol-grad}) requires exact knowledge of the
Q-values.  In practice they can be estimated, \eg, by the sum of discounted
rewards along an observed trajectory \citep{williams1992simple}, and the policy
gradient will still perform well \citep{konda2003onactor}.

\section{Regularized policy gradient algorithm}
\label{s-fixed-point}
In this section we derive a relationship between the policy and the Q-values
when using a regularized policy gradient algorithm. This allows us to transform
a policy into an estimate of the Q-values. We then show that for small
regularization the Q-values induced by the policy at the fixed point of the
algorithm have a small Bellman error in the tabular case.

\subsection{Tabular case}

Consider the fixed points of the entropy regularized policy gradient update
(\ref{e-rel-pol-grad}).
Let us define $f(\theta) = \Expect_{s,a} Q^\pi(s, a) \nabla_\theta \log\pi(s,a)
+ \alpha \Expect_s \nabla_\theta H(\pi_{s})$, and $g_s(\pi) = \sum_a \pi(s,a)$
for each $s$. A fixed point is one where we can no longer update $\theta$ in the
direction of $f(\theta)$ without violating one of the constraints $g_s(\pi) =
1$, \ie, where $f(\theta)$ is in the span of the vectors $\{\nabla_\theta
g_s(\pi)\}$. In other words, any fixed point must satisfy $f(\theta) = \sum_s
\lambda_s \nabla_\theta g_s(\pi)$, where for each $s$ the Lagrange multiplier
$\lambda_s \in \reals$ ensures that $g_s(\pi) = 1$. Substituting
in terms to this equation we obtain
\begin{equation}
  \label{e-fixed-point}
  \Expect_{s,a} \left(Q^\pi(s,a) - \alpha \log \pi(s,a) - c_s
  \right) \nabla_\theta \log \pi(s,a) = 0,
\end{equation}
where we have absorbed all constants into $c \in \reals^{|\X|}$.
Any solution $\pi$ to this equation is strictly positive element-wise
since it must lie in the domain of the entropy function.
In the tabular case $\pi$ is represented by a single number for
each state and action pair and the gradient
of the policy with respect to the parameters is the indicator function, \ie,
$\nabla_{\theta(t,b)} \pi(s,a) = \ones_{(t,b) = (s,a)}$. From this we obtain
$Q^\pi(s,a) - \alpha \log \pi(s,a) - c_s = 0$ for each $s$ (assuming that the
measure $d^\pi(s)>0$).  Multiplying by $\pi(a, s)$ and summing over $a \in \A$
we get $c_s = \alpha H^\pi(s) + V^\pi(s)$.
Substituting $c$ into equation (\ref{e-fixed-point}) we have the following
formulation for the policy: \begin{equation}
  \label{e-adv-policy}
  \pi(s, a) = \exp(A^\pi(s, a) / \alpha - H^\pi(s)),
\end{equation}
for all $s \in \X$ and $a \in \A$.
In other words, the policy at the fixed point is a softmax over the
advantage function induced by that policy, where the regularization parameter
$\alpha$ can be interpreted as the temperature. Therefore, we can
use the policy to derive an estimate of the Q-values,
\begin{equation}
\label{e-tilde-q1}
\tilde Q^\pi(s, a) = \tilde A^\pi(s, a) + V^\pi(s) = \alpha(\log \pi(s, a) +
H^\pi(s)) + V^\pi(s).
\end{equation}
With this we can rewrite the gradient update (\ref{e-rel-pol-grad}) as
\begin{equation}
\label{e-q-reg}
\Delta \theta \propto \Expect_{s,a}(Q^\pi(s,a) - \tilde Q^\pi(s,a))
\nabla_\theta \log \pi(s,a),
\end{equation}
since the update is unchanged by per-state constant offsets.
When the policy is parameterized as a softmax, \ie,
$\pi(s,a) = \exp(W(s,a)) / \sum_b \exp W(s,b) $, the quantity $W$ is sometimes
referred to as the action-preferences of the policy \cite[Chapter
6.6]{sutton:book}. Equation (\ref{e-adv-policy}) states that the action
preferences are equal to the Q-values scaled by $1/\alpha$, up to an additive
per-state constant.

\subsection{General case}
Consider the following optimization problem:
\begin{equation}
\label{e-opt-regression}
\begin{array}{ll}
  \mbox{minimize} & \Expect_{s,a} (q(s,a) - \alpha \log \pi(s,a))^2 \\
  \mbox{subject to} & \sum_a \pi(s, a) = 1, \quad s \in \X
\end{array}
\end{equation}
over variable $\theta$ which parameterizes $\pi$, where we consider both the measure in the
expectation and the values $q(s,a)$ to be independent of $\theta$. The
optimality condition for this problem is
\[
  \Expect_{s,a}(q(s,a) - \alpha \log \pi(s,a) + c_s) \nabla_\theta \log \pi(s,a) = 0,
\]
where $c \in \reals^{|\X|}$ is the Lagrange multiplier associated with the
constraint that the policy sum to one at each state.  Comparing this to equation
(\ref{e-fixed-point}), we see that if $q = Q^\pi$ and the measure in the
expectation is the same then they describe the same set of fixed points.  This
suggests an interpretation of the fixed points of the regularized policy
gradient as a regression of the log-policy onto the Q-values. In the general
case of using an approximation architecture we can interpret equation
(\ref{e-fixed-point}) as indicating that the error between $Q^\pi$ and $\tilde
Q^\pi$ is orthogonal to $\nabla_{\theta_i} \log \pi$ for each $i$, and so cannot
be reduced further by changing the parameters, at least locally.  In this case
equation (\ref{e-adv-policy}) is unlikely to hold at a solution to
(\ref{e-fixed-point}), however with a good approximation architecture it may
hold approximately, so that the we can derive an \emph{estimate} of the Q-values
from the policy using equation (\ref{e-tilde-q1}). We will use this estimate of
the Q-values in the next section.

\subsection{Connection to action-value methods}
The previous section made a connection between regularized policy gradient and
a regression onto the Q-values at the fixed point. In this section we go one
step further, showing that actor-critic methods can be interpreted as
action-value fitting methods, where the exact method depends on the choice of
critic.

\paragraph{Actor-critic methods.}
Consider an agent using an actor-critic method to learn both a policy $\pi$ and
a value function $V$. At any iteration $k$, the value function $V^k$ has
parameters $w^k$, and the policy is of the form
\begin{equation}
  \label{e-con-pol1}
  \pi^k(s,a) = \exp(W^k(s,a) / \alpha)  / \sum_b \exp(W^k(s,b) / \alpha),
\end{equation}
where $W^k$ is parameterized by $\theta^k$ and $\alpha > 0$ is the entropy
regularization penalty. In this case
$\nabla_\theta \log \pi^k(s,a) = (1/\alpha)(\nabla_\theta W^k(s,a) - \sum_b
\pi(s,b) \nabla_\theta W^k(s,b))$.
Using equation (\ref{e-q-reg})
the parameters are updated as
\begin{equation}
  \label{e-p-update}
  \Delta \theta \propto \Expect_{s,a} \delta_\mathrm{ac}(\nabla_\theta W^k(s,a) -
  \sum_b \pi^k(s,b) \nabla_\theta  W^k(s,b)), \quad
  \Delta w \propto \Expect_{s,a} \delta_\mathrm{ac} \nabla_w V^k(s)
\end{equation}
where $\delta_\mathrm{ac}$ is the \emph{critic minus baseline} term, which depends
on the variant of actor-critic being used (see the remark below).

\paragraph{Action-value methods.}
Compare this to the case where an agent is learning Q-values with a dueling
architecture \citep{wang2015dueling}, which at iteration $k$ is given by
\[
Q^k(s,a) = Y^k(s,a) - \sum_b \mu(s,b) Y^k(s,b) + V^k(s),
\]
where $\mu$
is a probability distribution, $Y^k$ is parameterized by $\theta^k$, $V^k$
is parameterized by $w^k$, and the exploration policy is Boltzmann with
temperature $\alpha$, \ie,
\begin{equation}
  \label{e-con-pol2}
  \pi^k(s,a) = \exp(Y^k(s,a) / \alpha)  / \sum_b \exp(Y^k(s,b) / \alpha).
\end{equation}
In action value fitting methods at each iteration the parameters are updated to
reduce some error, where the update is given by
\begin{equation}
  \label{e-q-update}
  \Delta \theta \propto \Expect_{s,a} \delta_\mathrm{av}(\nabla_\theta Y^k(s,a) - \sum_b \mu(s,b) \nabla_\theta
  Y^k(s,b)), \quad
  \Delta w \propto \Expect_{s,a} \delta_\mathrm{av} \nabla_w V^k(s)
\end{equation}
where $\delta_\mathrm{av}$ is the \emph{action-value error} term and depends on
which algorithm is being used (see the remark below).

\paragraph{Equivalence.}
The two policies (\ref{e-con-pol1}) and (\ref{e-con-pol2}) are identical if $W^k
= Y^k$ for all $k$. Since $X^0$ and $Y^0$ can be initialized and parameterized
in the same way, and assuming the two value function estimates
are initialized and parameterized in the same way, all that remains is to show
that the updates in equations (\ref{e-q-update}) and (\ref{e-p-update}) are
identical.  Comparing the two, and assuming that $\delta_\mathrm{ac} =
\delta_\mathrm{av}$ (see remark), we see that the only difference is that the
measure is not fixed in (\ref{e-p-update}), but is equal to the current
policy and therefore changes after each update. Replacing $\mu$ in
(\ref{e-q-update}) with $\pi^k$
makes the updates identical, in which case $W^k = Y^k$ at all iterations and the
two policies (\ref{e-con-pol1}) and (\ref{e-con-pol2}) are always the same. In
other words, the slightly modified action-value method is equivalent to an actor-critic
policy gradient method,
and vice-versa (modulo using the non-discounted distribution of states, as
discussed in \S \ref{s-pg}). In particular, regularized policy gradient methods
can be interpreted as advantage function learning techniques
\citep{baird1993advantage}, since at the optimum the quantity $W(s,a) - \sum_b
\pi(s,b) W(s,b) = \alpha(\log \pi(s,a) + H^\pi(s))$ will be equal to the
advantage function values in the tabular case.

\paragraph{Remark.}
In SARSA \citep{rummery1994sarsa} we set $\delta_\mathrm{av} = r(s,a) + \gamma
Q(s^\prime ,b) - Q(s,a)$, where $b$ is the action selected at state $s^\prime$,
which would be equivalent to using a bootstrap critic in equation
(\ref{e-q-reg}) where $Q^\pi(s,a) = r(s,a) + \gamma \tilde Q(s^\prime, b)$.  In
expected-SARSA \citep[Exercise 6.10]{sutton:book},
\citep{seijen2009expectedsarsa}) we take the expectation over the Q-values at
the next state, so $\delta_\mathrm{av} = r(s,a) + \gamma V(s^\prime) - Q(s,a)$.
This is equivalent to TD-actor-critic \citep{konda2003onactor} where we use the
value function to provide the critic, which is given by $Q^\pi = r(s,a) + \gamma
V(s^\prime)$.  In Q-learning \citep{watkins1989learning} $\delta_\mathrm{av} =
r(s,a) + \gamma \max_b Q(s^\prime, b) - Q(s,a)$, which would be equivalent to
using an optimizing critic that bootstraps using the max Q-value at the next
state, \ie, $Q^\pi(s,a) = r(s,a) + \gamma \max_b \tilde Q^\pi(s^\prime, b)$. In
REINFORCE the critic is the Monte Carlo return from that state on, \ie,
$Q^\pi(s,a) = (\sum_{t=0}^\infty \gamma^t r_t \mid s_0 = s, a_0 = a)$. If the
return trace is truncated and a bootstrap is performed after $n$-steps, this is
equivalent to $n$-step SARSA or $n$-step Q-learning, depending on the form of
the bootstrap \citep{peng1996msq}.

\subsection{Bellman residual}
\label{s-bell-err}
In this section we show that $\|\T^\star Q^{\pi_\alpha} - Q^{\pi_\alpha} \|
\rightarrow 0$ with decreasing regularization penalty $\alpha$, where
$\pi_\alpha$ is the policy defined by $(\ref{e-adv-policy})$ and
$Q^{\pi_\alpha}$ is the corresponding Q-value function, both of which are
functions of $\alpha$.  We shall show that it converges to zero by bounding the
sequence below by zero and above with a sequence that converges to zero. First,
we have that $\T^\star Q^{\pi_\alpha} \geq \T^{\pi_\alpha} Q^{\pi_\alpha} =
Q^{\pi_\alpha}$, since $\T^\star$ is greedy with respect to the Q-values.  So
$\T^\star Q^{\pi_\alpha} - Q^{\pi_\alpha} \geq 0$. Now, to bound from above we
need the fact that $\pi_\alpha(s,a)= \exp(Q^{\pi_\alpha}(s,a) / \alpha) / \sum_b
\exp(Q^{\pi_\alpha}(s,b) / \alpha) \leq \exp((Q^{\pi_\alpha}(s,a) - \max_c
Q^{\pi_\alpha}(s,c)) / \alpha)$. Using this we have
\[
\begin{array}{rcl}
  0 &\leq& \T^\star Q^{\pi_\alpha}(s,a) - Q^{\pi_\alpha}(s,a)\\
         &=& \T^\star Q^{\pi_\alpha}(s,a) - \T^{\pi_\alpha} Q^{\pi_\alpha}(s,a) \\
  &=& \Expect_{s^\prime}(\max_c Q^{\pi_\alpha}(s^\prime, c) - \sum_b \pi_\alpha(s^\prime, b) Q^{\pi_\alpha}(s^\prime, b)) \\
    &=& \Expect_{s^\prime} \sum_b \pi_\alpha(s^\prime, b) (\max_c Q^{\pi_\alpha}(s^\prime, c) - Q^{\pi_\alpha}(s^\prime, b)) \\
                                                       &\leq& \Expect_{s^\prime}
    \sum_b \exp((Q^{\pi_\alpha}(s^\prime,b) - Q^{\pi_\alpha}(s^\prime,
    b^\star)) / \alpha)(\max_c Q^{\pi_\alpha}(s^\prime, c) - Q^{\pi_\alpha}(s^\prime, b)) \\
&=& \Expect_{s^\prime} \sum_b f_\alpha (\max_c Q^{\pi_\alpha}(s^\prime,c) - Q^{\pi_\alpha}(s^\prime,
b)),
  \end{array}
\]
where we define $f_\alpha (x) = x \exp(-x / \alpha)$. To conclude our proof
we use the fact that $f_\alpha(x) \leq \sup_x f_\alpha(x) = f_\alpha(\alpha) =
\alpha \mathrm{e}^{-1}$, which yields
\[
  0 \leq \T^\star Q^{\pi_\alpha}(s,a) - Q^{\pi_\alpha}(s,a) \leq |\A|\alpha \mathrm{e}^{-1}
\]
for all $(s,a)$, and so the Bellman residual converges to zero with
decreasing $\alpha$.
In other words, for small enough $\alpha$ (which is the regime we are interested
in) the Q-values induced by the policy (\ref{e-adv-policy}) will have a small
Bellman residual. Moreover, this implies that $\lim_{\alpha \rightarrow 0}
Q^{\pi_\alpha} = Q^\star$, as one might expect.

\section{PGQL}
In this section we introduce the main contribution of the paper, which is a
technique to combine policy gradient with Q-learning. We call our technique
`PGQL', for policy gradient and Q-learning.  In the
previous section we showed that the Bellman residual is small at the fixed point
of a regularized policy gradient algorithm when the regularization penalty is
sufficiently small.  This suggests adding an auxiliary update where we
explicitly attempt to reduce the Bellman residual as estimated from the policy,
\ie, a hybrid between policy gradient and Q-learning.

We first present the technique in a batch update setting, with a perfect
knowledge of $Q^\pi$ (\ie, a perfect critic).  Later we discuss the practical
implementation of the technique in a reinforcement learning setting with
function approximation, where the agent generates experience from interacting
with the environment and needs to estimate a critic simultaneously with the
policy.

\subsection{PGQL update}
Define the estimate of $Q$ using the policy as
\begin{equation}
  \label{e-q-est}
  \tilde Q^\pi(s, a) = \alpha(\log \pi(s, a) + H^\pi(s)) + V(s),
\end{equation}
where $V$ has parameters $w$ and is not necessarily $V^\pi$ as it was in
equation (\ref{e-tilde-q1}). In (\ref{e-rel-pol-grad}) it
was unnecessary to estimate the constant since the update was invariant to
constant offsets, although in practice it is often estimated for use in a
variance reduction technique \citep{williams1992simple, sutton1999policy}.

Since we know that at the fixed point the Bellman residual will be small for small
$\alpha$, we can
consider updating the parameters to reduce the Bellman residual in a fashion
similar to Q-learning, \ie,
\begin{equation}
  \label{e-q-learn}
  \Delta \theta \propto \Expect_{s,a} (\T^\star \tilde
  Q^\pi(s,a) - \tilde Q^\pi(s,a)) \nabla_\theta \log \pi(s,a), \quad
  \Delta w \propto \Expect_{s,a} (\T^\star \tilde Q^\pi(s,a) - \tilde Q^\pi(s,a)) \nabla_w
  V(s).
\end{equation}
This is Q-learning applied to a particular form of the Q-values, and can also be
interpreted as an actor-critic algorithm with an optimizing (and therefore
biased) critic.

The full scheme simply combines two updates to the policy, the
regularized policy gradient update (\ref{e-rel-pol-grad}) and the Q-learning
update (\ref{e-q-learn}).
Assuming we have an architecture that provides a policy $\pi$, a value function
estimate $V$, and an action-value critic $Q^\pi$, then the parameter updates can be
written as (suppressing the $(s,a)$ notation)
\begin{equation}
\label{e-hybrid}
\begin{array}{c}
\Delta \theta \propto  (1-\eta)\Expect_{s,a} (Q^\pi - \tilde Q^\pi) \nabla_\theta
\log\pi+\eta \Expect_{s,a} (\T^\star \tilde Q^\pi - \tilde Q^\pi)
\nabla_\theta\log\pi, \\ \\
\Delta w \propto  (1-\eta)\Expect_{s,a} (Q^\pi - \tilde Q^\pi) \nabla_w
V+\eta \Expect_{s,a} (\T^\star \tilde Q^\pi - \tilde Q^\pi) \nabla_w V,
\end{array}
\end{equation}
here $\eta \in [0,1]$ is a weighting parameter that controls how much of each
update we apply. In the case where $\eta = 0$ the above scheme reduces to
entropy regularized policy gradient. If $\eta = 1$ then it becomes a variant of
(batch) Q-learning with an architecture similar to the dueling architecture
\citep{wang2015dueling}.  Intermediate values of $\eta$ produce a hybrid between
the two.  Examining the
update we see that two error terms are trading off. The first term encourages
consistency with critic, and the second term encourages optimality over time.
However, since we know that under standard policy gradient the Bellman residual
will be small, then it follows that adding a term that reduces that error should
not make much difference at the fixed point. That is, the updates should be
complementary, pointing in the same general direction, at least far away from a
fixed point.  This update can also be interpreted as an actor-critic update
where the critic is given by a weighted combination of a standard critic and an
optimizing critic. Yet another interpretation of the update is a combination of
expected-SARSA and Q-learning, where the Q-values are parameterized as the sum
of an advantage function and a value function.

\subsection{Practical implementation}

The updates presented in (\ref{e-hybrid}) are batch
updates, with an exact critic $Q^\pi$. In practice we want to run this scheme
online, with an estimate of the critic, where we don't necessarily apply the
policy gradient update at the same time or from same data source as the
Q-learning update.

Our proposal scheme is as follows. One or more agents interact with an
environment, encountering states and rewards and performing on-policy updates of
(shared) parameters using an actor-critic algorithm where both the policy and
the critic are being updated online. Each time an agent receives new data from
the environment it writes it to a shared replay memory buffer.  Periodically a
separate learner process samples from the replay buffer and performs a step of
Q-learning on the parameters of the policy using (\ref{e-q-learn}).  This scheme
has several advantages.  The critic can accumulate the Monte Carlo return over
many time periods, allowing us to spread the influence of a reward received in
the future backwards in time. Furthermore, the replay buffer can be used to
store and replay `important' past experiences by prioritizing those samples
\citep{schaul2015prioritized}. The use of the replay buffer can help to reduce
problems associated with correlated training data, as generated by an agent
exploring an environment where the states are likely to be similar from one time
step to the next.  Also the use of replay can act as a kind of regularizer,
preventing the policy from moving too far from satisfying the Bellman equation,
thereby improving stability, in a similar sense to that of a policy
`trust-region' \citep{schulman2015trust}. Moreover, by batching up replay
samples to update the network we can leverage GPUs to perform the updates
quickly, this is in comparison to pure policy gradient techniques which are
generally implemented on CPU \citep{mnih2016asynchronous}.

Since we perform Q-learning using samples from a replay buffer that were
generated by a old policy we are performing (slightly) off-policy learning.
However, Q-learning is known to converge to the optimal Q-values in the
off-policy tabular case (under certain conditions) \citep{sutton:book}, and
has shown good performance off-policy in the function approximation case
\citep{mnih-atari-2013}.

\subsection{Modified fixed point}
The PGQL updates in equation (\ref{e-hybrid}) have
modified the fixed point of the algorithm, so the analysis of \S
\ref{s-fixed-point} is no longer valid. Considering the tabular case once again,
it is still the case that the policy
$\pi \propto \exp(\tilde Q^\pi / \alpha)$ as before, where $\tilde Q^\pi$ is
defined by (\ref{e-q-est}), however where previously
the fixed point satisfied $\tilde Q^\pi = Q^\pi$, with $Q^\pi$ corresponding
to the Q-values induced by $\pi$, now we have
\begin{equation}
\label{e-mod-fixed-pt}
\tilde Q^\pi = (1-\eta) Q^\pi + \eta \T^\star \tilde Q^\pi,
\end{equation}
Or equivalently, if $\eta < 1$, we have
$\tilde Q^\pi = (1 - \eta) \sum_{k=0}^\infty \eta^k (\T^\star)^k Q^\pi$.
In the appendix we show that $\|\tilde Q^\pi - Q^\pi \| \rightarrow 0$ and that
$\|\T^\star Q^\pi - Q^\pi \| \rightarrow 0$ with decreasing $\alpha$ in the
tabular case. That is, for small $\alpha$ the induced Q-values and the Q-values
estimated from the policy are close, and we still have the guarantee that
in the limit the Q-values are optimal. In other words, we
have not perturbed the policy very much by the addition of the auxiliary update.


\section{Numerical experiments}

\subsection{Grid world}
\begin{figure}
\begin{center}
  \makebox[\linewidth][c]{
    \centering
    \begin{subfigure}[b]{0.3\textwidth}
        \includegraphics[width=\textwidth]{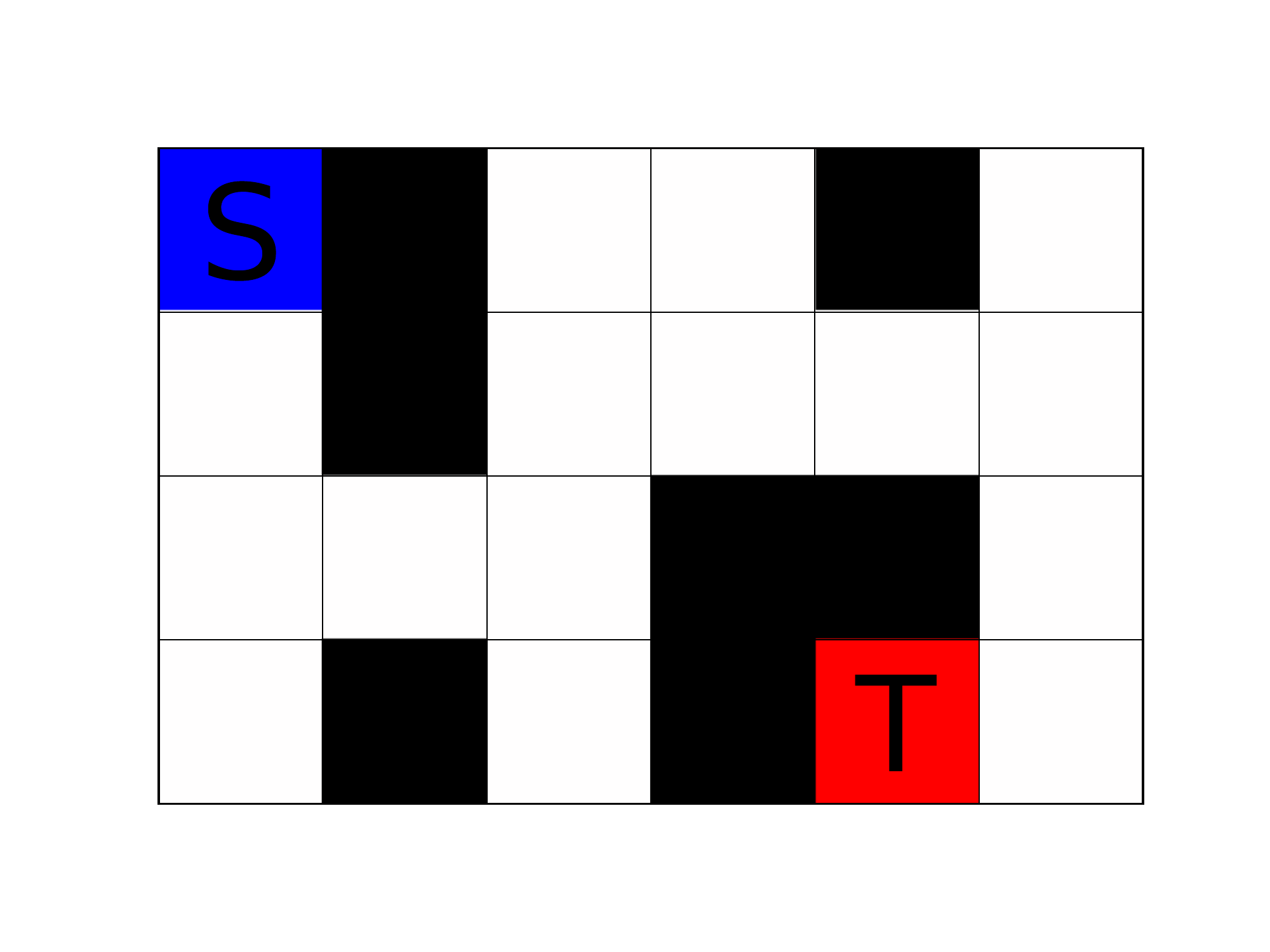}
          \caption{Grid world.}
          \label{f-grid-world}
    \end{subfigure}
    \begin{subfigure}[b]{0.5\textwidth}
        \includegraphics[width=\textwidth]{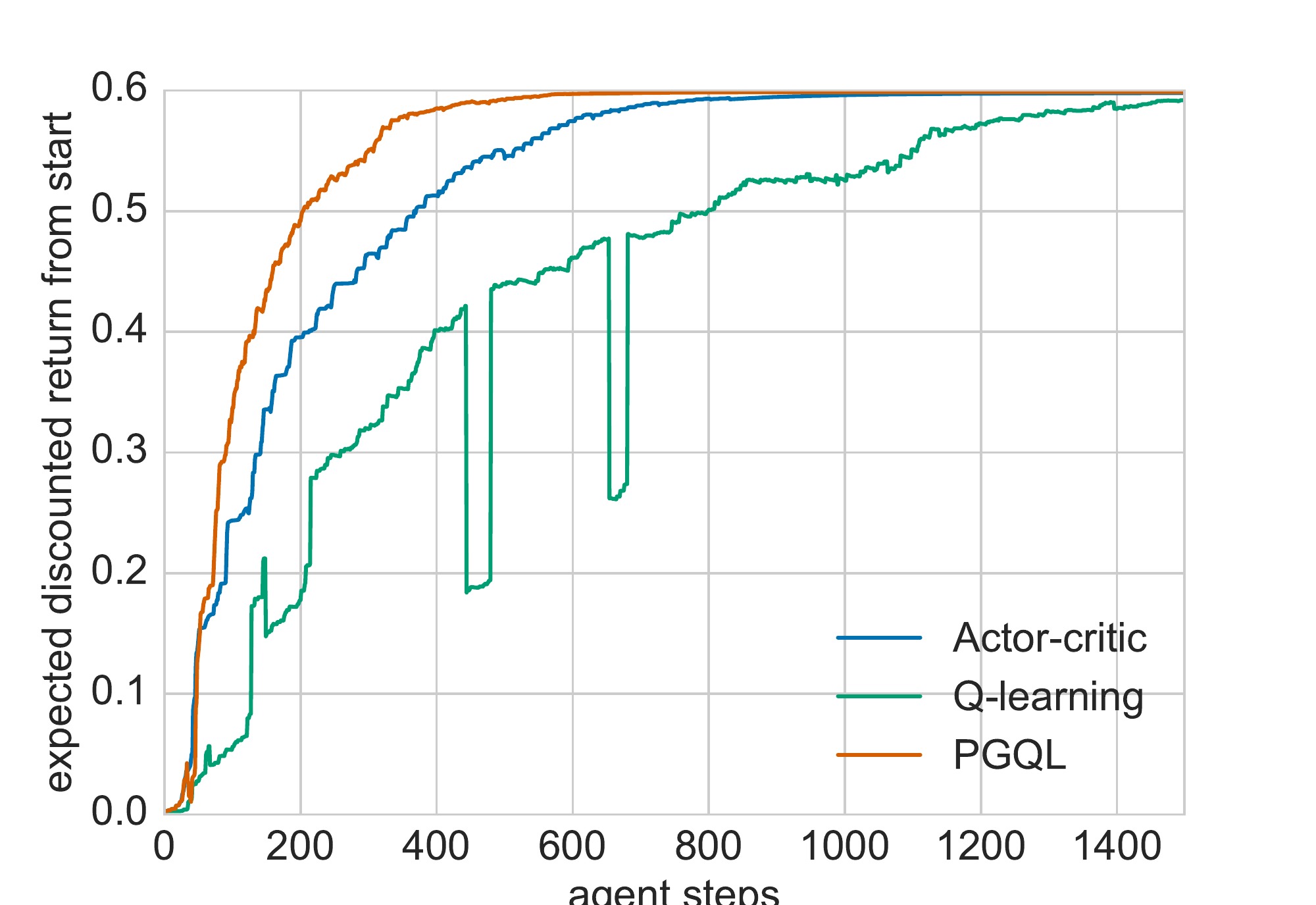}
          \caption{Performance versus agent steps in grid world.}
          \label{f-gw-scores}
    \end{subfigure}
  }
  \caption{Grid world experiment.
  \label{f-grid-world-full}
  }
\end{center}
\end{figure}

In this section we discuss the results of running PGQL on a toy $4$ by
$6$ grid world, as shown in Figure \ref{f-grid-world}. The agent always begins
in the square marked `S' and the episode continues until it reaches the square
marked `T', upon which it receives a reward of $1$. All other times it receives
no reward. For this experiment we chose regularization parameter $\alpha = 0.001$
and discount factor $\gamma = 0.95$.

Figure \ref{f-gw-scores} shows the performance traces of three different agents
learning in the grid world, running from the same initial random seed. The lines
show the \emph{true} expected performance of the policy from the start state, as
calculated by value iteration after each update.  The blue-line is standard
TD-actor-critic \citep{konda2003onactor}, where we maintain an estimate of the
value function and use that to generate an estimate of the Q-values for use as
the critic. The green line is Q-learning where at each step an update is
performed using data drawn from a replay buffer of prior experience and where
the Q-values are parameterized as in equation (\ref{e-q-est}). The policy is a
softmax over the Q-value estimates with temperature $\alpha$. The red line is
PGQL, which at each step first performs the TD-actor-critic update, then performs
the Q-learning update as in (\ref{e-hybrid}).

The grid world was totally deterministic, so the step size could be large and
was chosen to be $1$. A step-size any larger than this made the pure
actor-critic agent fail to learn, but both PGQL and Q-learning could handle
some increase in the step-size, possibly due to the stabilizing effect of using
replay.

It is clear that PGQL outperforms the other two. At any point along the
x-axis the agents have seen the same amount of data, which would indicate that
PGQL is more data efficient than either of the vanilla methods since
it has the highest performance at practically every point.

\subsection{Atari}

\begin{figure}
\begin{center}
  \includegraphics[scale=.3]{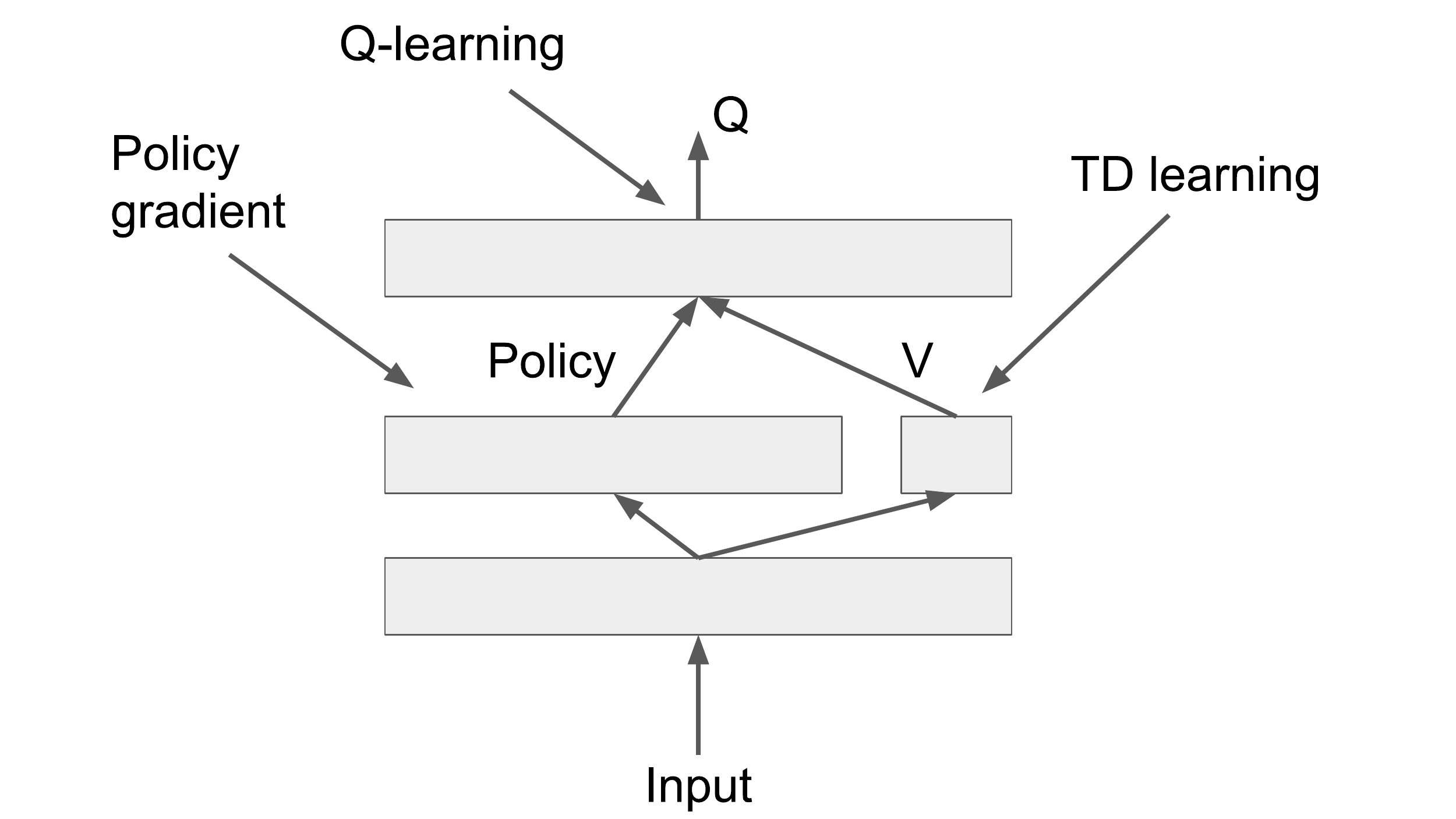}
  \caption{PGQL network augmentation.
  \label{f-net}
  }
\end{center}
\end{figure}

We tested our algorithm on the full suite of Atari benchmarks
\citep{bellemare-ale}, using a neural network to parameterize the policy.
In figure \ref{f-net} we show how a policy network can be augmented with a
parameterless additional layer which outputs the Q-value estimate.
With the exception of the extra layer, the architecture
and parameters were chosen to exactly match the asynchronous advantage
actor-critic (A3C) algorithm presented in \citet{mnih2016asynchronous}, which in
turn reused many of the settings from \citet{mnih-dqn-2015}. Specifically we
used the exact same learning rate, number of workers, entropy penalty, bootstrap
horizon, and network architecture.  This allows a fair comparison between A3C
and PGQL, since the only difference is the addition of the
Q-learning step.  Our technique augmented A3C with the following change: After
each actor-learner has accumulated the gradient for the policy update, it
performs a single step of Q-learning from replay data as described in equation
(\ref{e-q-learn}), where the minibatch size was 32 and the Q-learning learning
rate was chosen to be $0.5$ times the actor-critic learning rate (we mention
learning rate ratios rather than choice of $\eta$ in (\ref{e-hybrid}) because the
updates happen at different frequencies and from different data sources). Each
actor-learner thread maintained a replay buffer of the last $100k$ transitions
seen by that thread.  We ran the learning for $50$ million agent steps ($200$
million Atari frames), as in \citep{mnih2016asynchronous}.

In the results we compare against both A3C and a variant of asynchronous deep
Q-learning.  The changes we made to Q-learning are to make it similar to our
method, with some tuning of the hyper-parameters for performance.
We use the exact same network, the exploration policy is a softmax over the
Q-values with a temperature of $0.1$, and the Q-values are parameterized as in
equation (\ref{e-q-est}) (\ie, similar to the dueling architecture
\citep{wang2015dueling}), where $\alpha = 0.1$.  The Q-value updates are
performed every 4 steps with a minibatch of 32 (roughly 5 times more frequently
than PGQL). For each method, all games used identical hyper-parameters.

The results across all games are given in table \ref{t-atari} in the appendix.
All scores have been normalized by subtracting the average score achieved by an
agent that takes actions uniformly at random.  Each game was tested 5 times per
method with the same hyper-parameters but with different random seeds.  The
scores presented correspond to the best score obtained by any run from a random
start evaluation condition \citep{mnih2016asynchronous}.  Overall, PGQL
performed best in 34 games, A3C performed best in 7 games, and Q-learning was
best in 10 games. In 6 games two or more methods tied. In tables
\ref{t-atari-means-random} and \ref{t-atari-means-human} we give the mean and
median normalized scores as percentage of an expert human normalized score
across all games for each tested algorithm from random and human-start
conditions respectively. In a human-start condition the agent takes over control
of the game from randomly selected human-play starting points, which generally
leads to lower performance since the agent may not have found itself in that
state during training.  In both cases, PGQL has both the highest mean and median,
and the median score exceeds 100\%, the human performance threshold.

\begin{figure}[h]
  \makebox[\linewidth][c]{
    \centering
    \begin{subfigure}[b]{0.4\textwidth}
        \includegraphics[width=\textwidth]{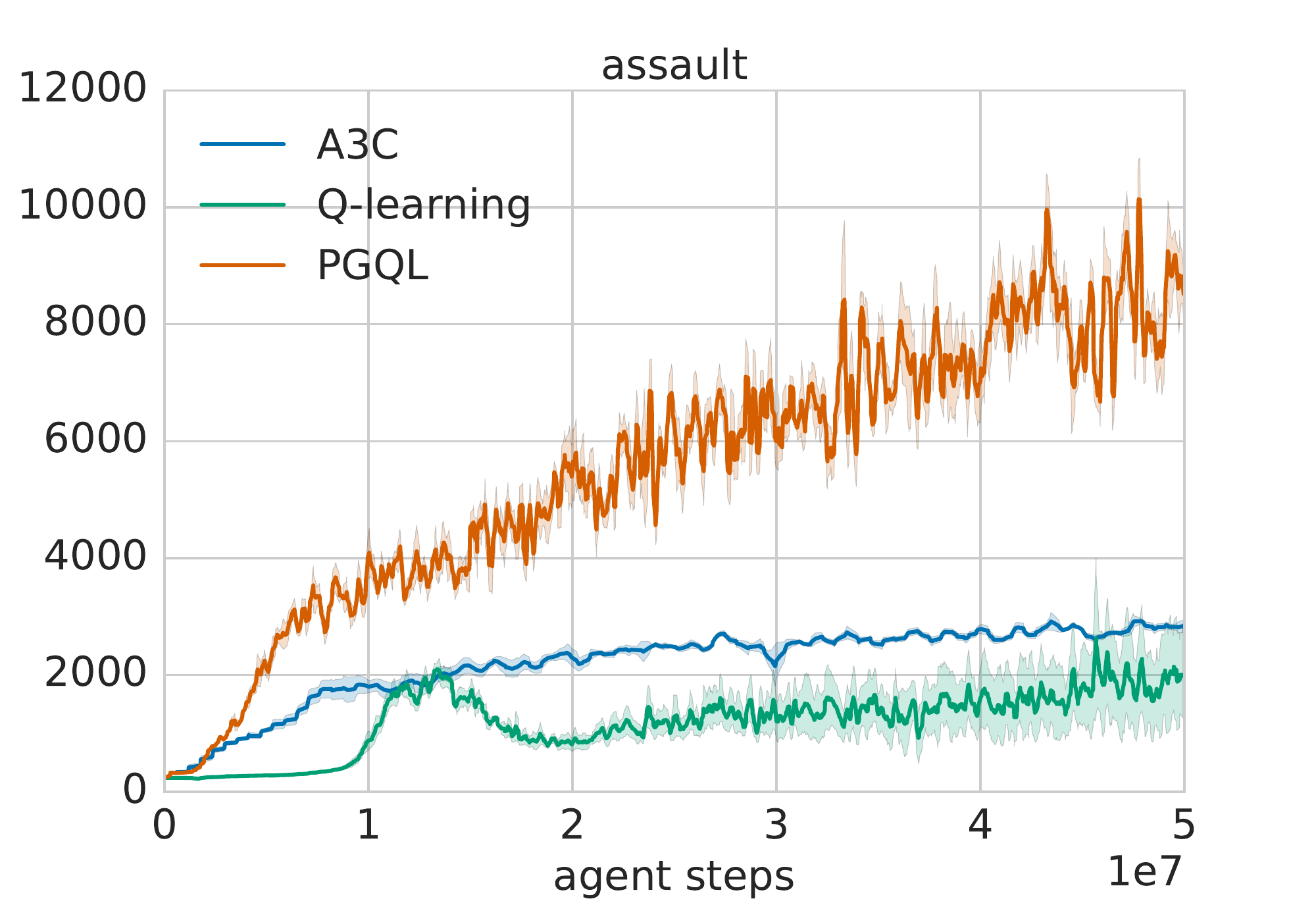}
    \end{subfigure}
    \begin{subfigure}[b]{0.4\textwidth}
        \includegraphics[width=\textwidth]{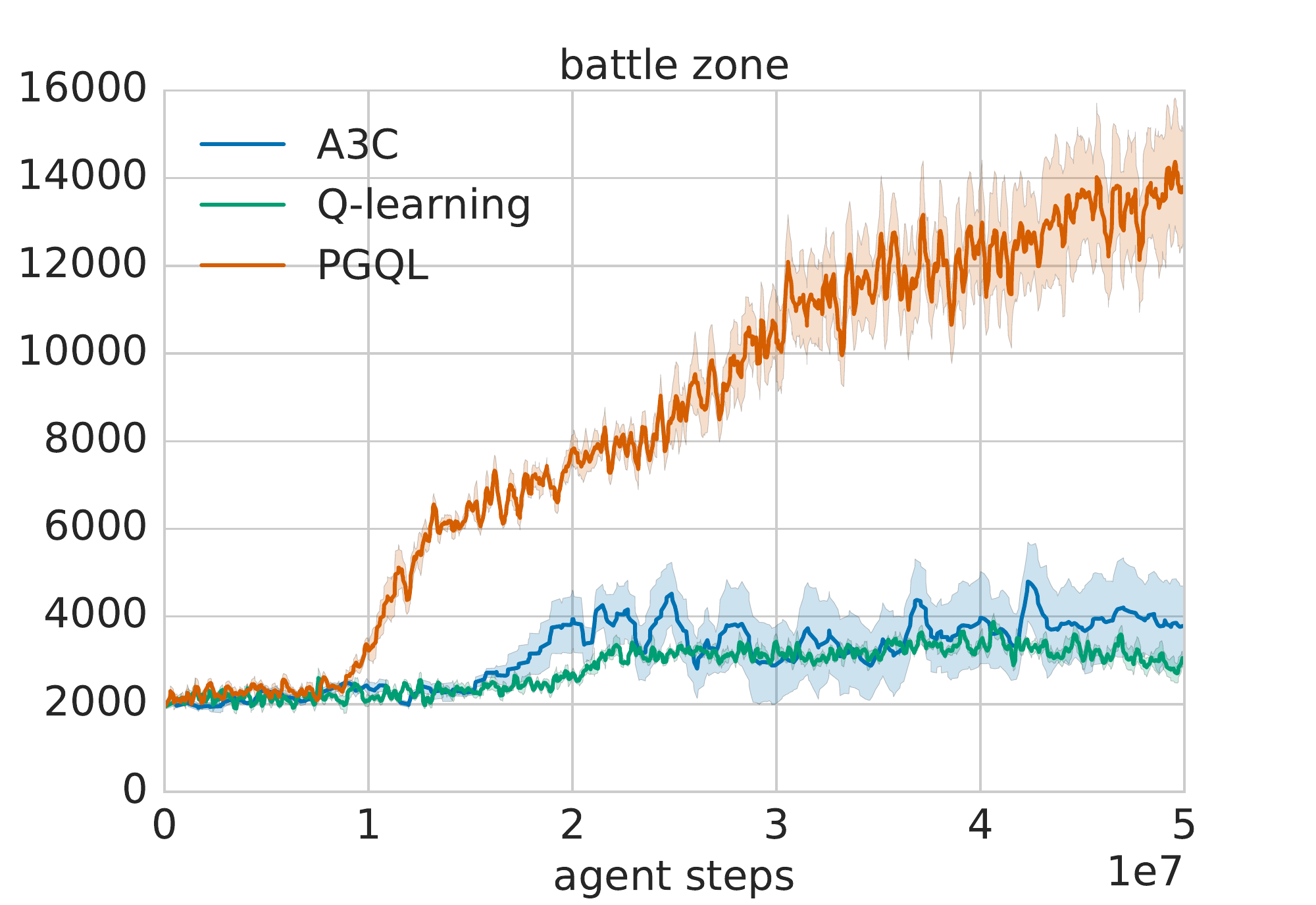}
    \end{subfigure}
  }
  \makebox[\linewidth][c]{
    \centering
    \begin{subfigure}[b]{0.4\textwidth}
        \includegraphics[width=\textwidth]{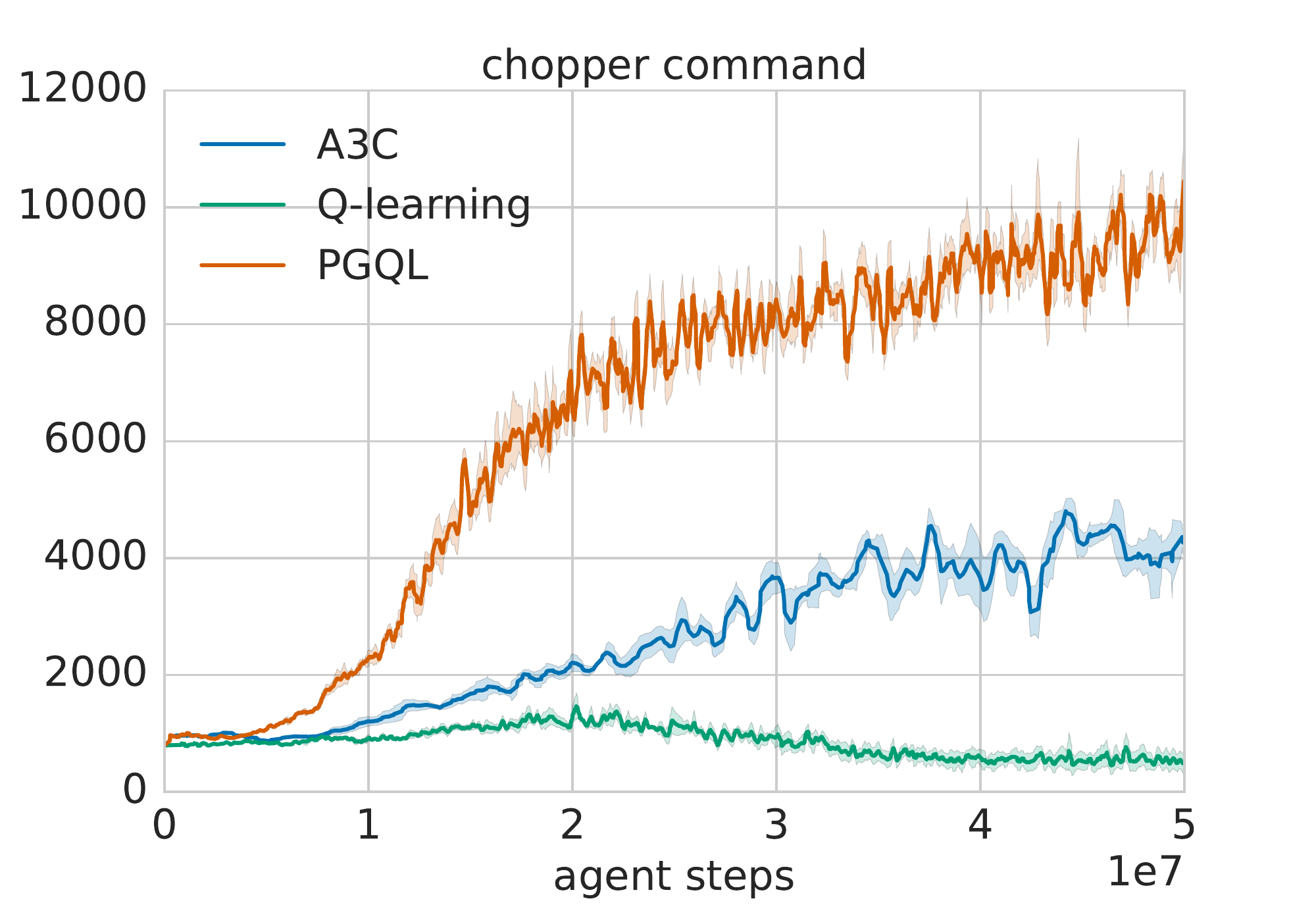}
    \end{subfigure}
    \begin{subfigure}[b]{0.4\textwidth}
        \includegraphics[width=\textwidth]{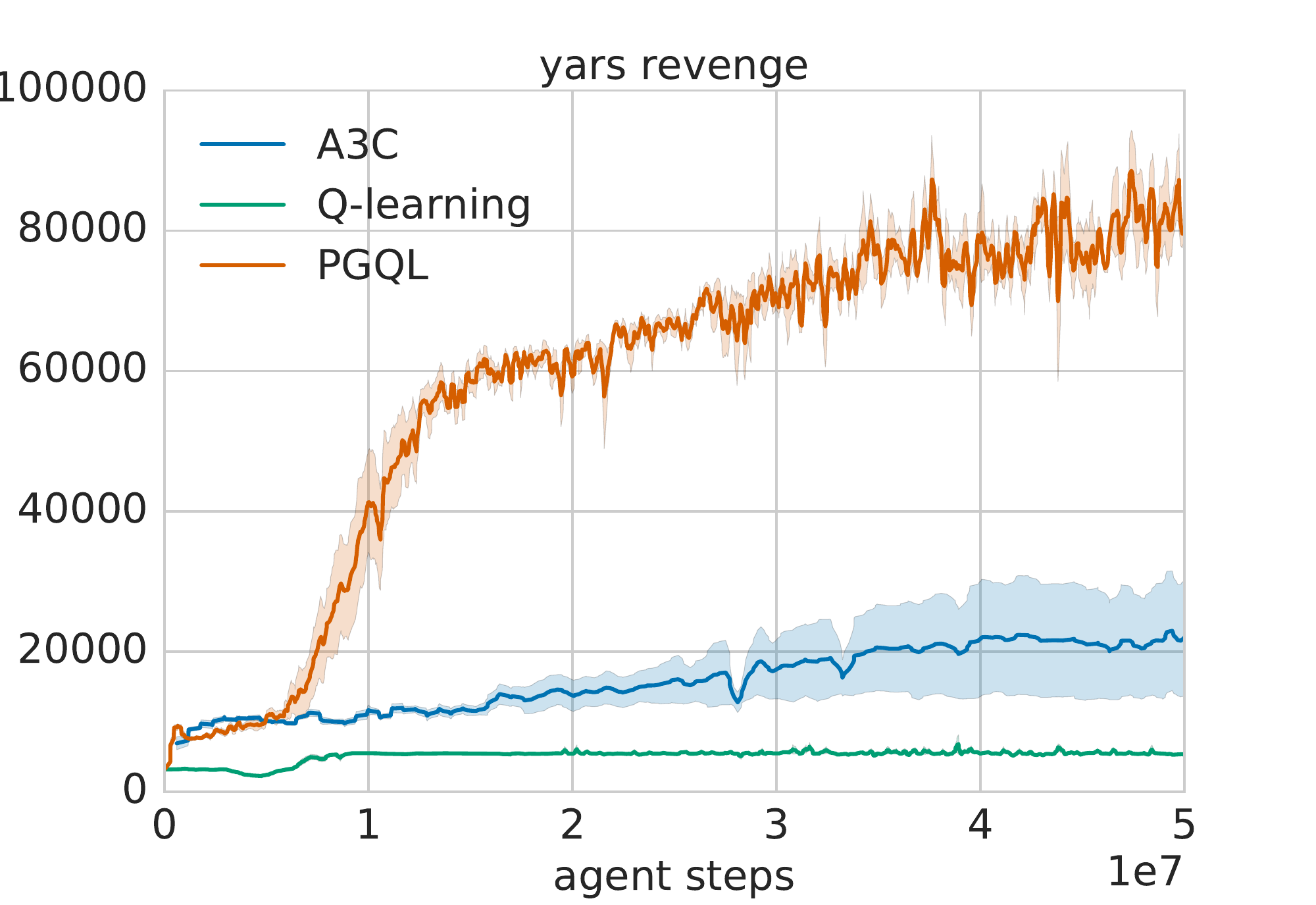}
    \end{subfigure}
  }
  \caption{Some Atari runs where PGQL performed well.
  \label{f-atari-successes}
  }
\end{figure}

\begin{figure}[h]
  \makebox[\linewidth][c]{
    \centering
    \begin{subfigure}[b]{0.4\textwidth}
        \includegraphics[width=\textwidth]{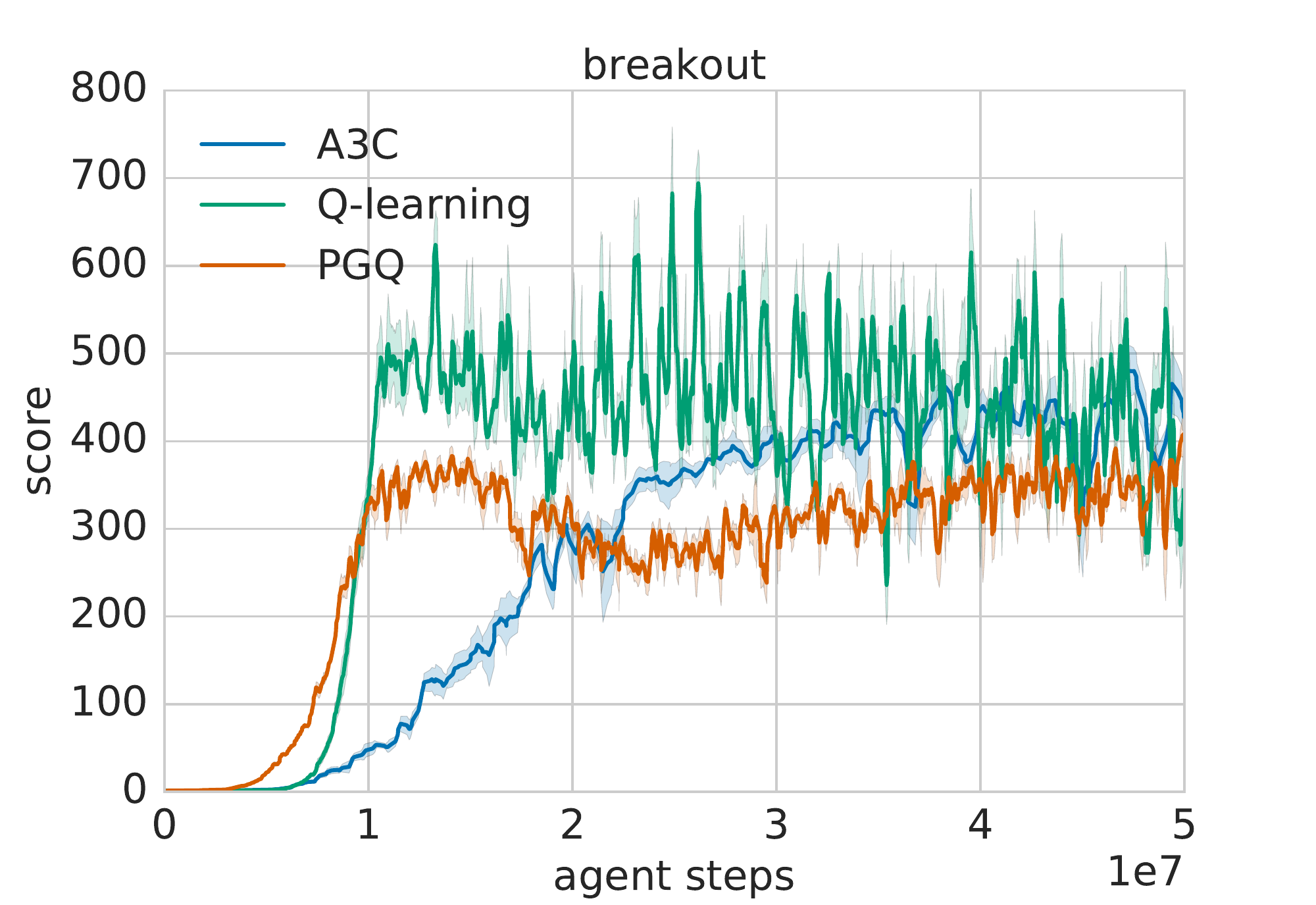}
    \end{subfigure}
    \begin{subfigure}[b]{0.4\textwidth}
        \includegraphics[width=\textwidth]{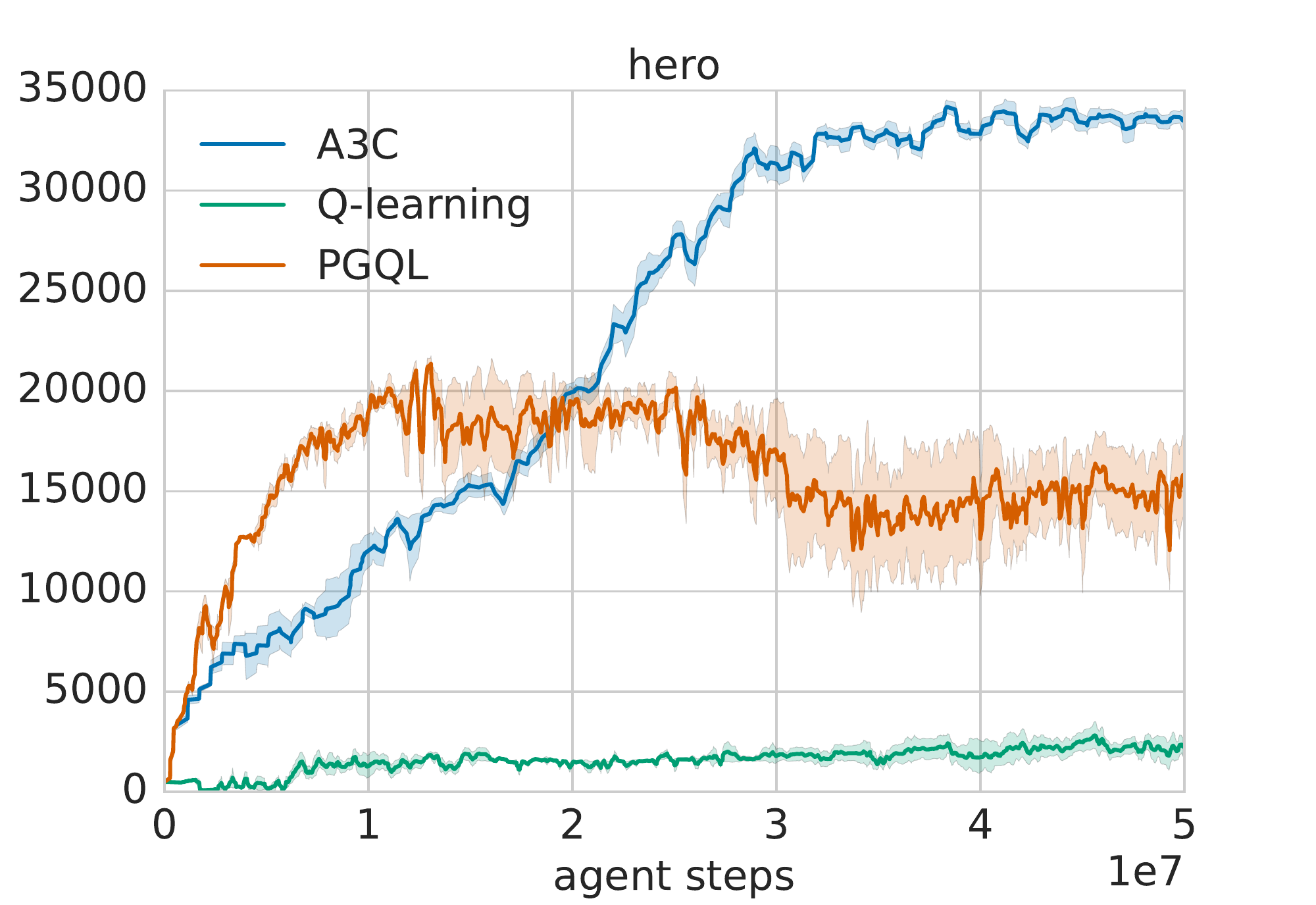}
    \end{subfigure}
  }
  \makebox[\linewidth][c]{
    \centering
    \begin{subfigure}[b]{0.4\textwidth}
        \includegraphics[width=\textwidth]{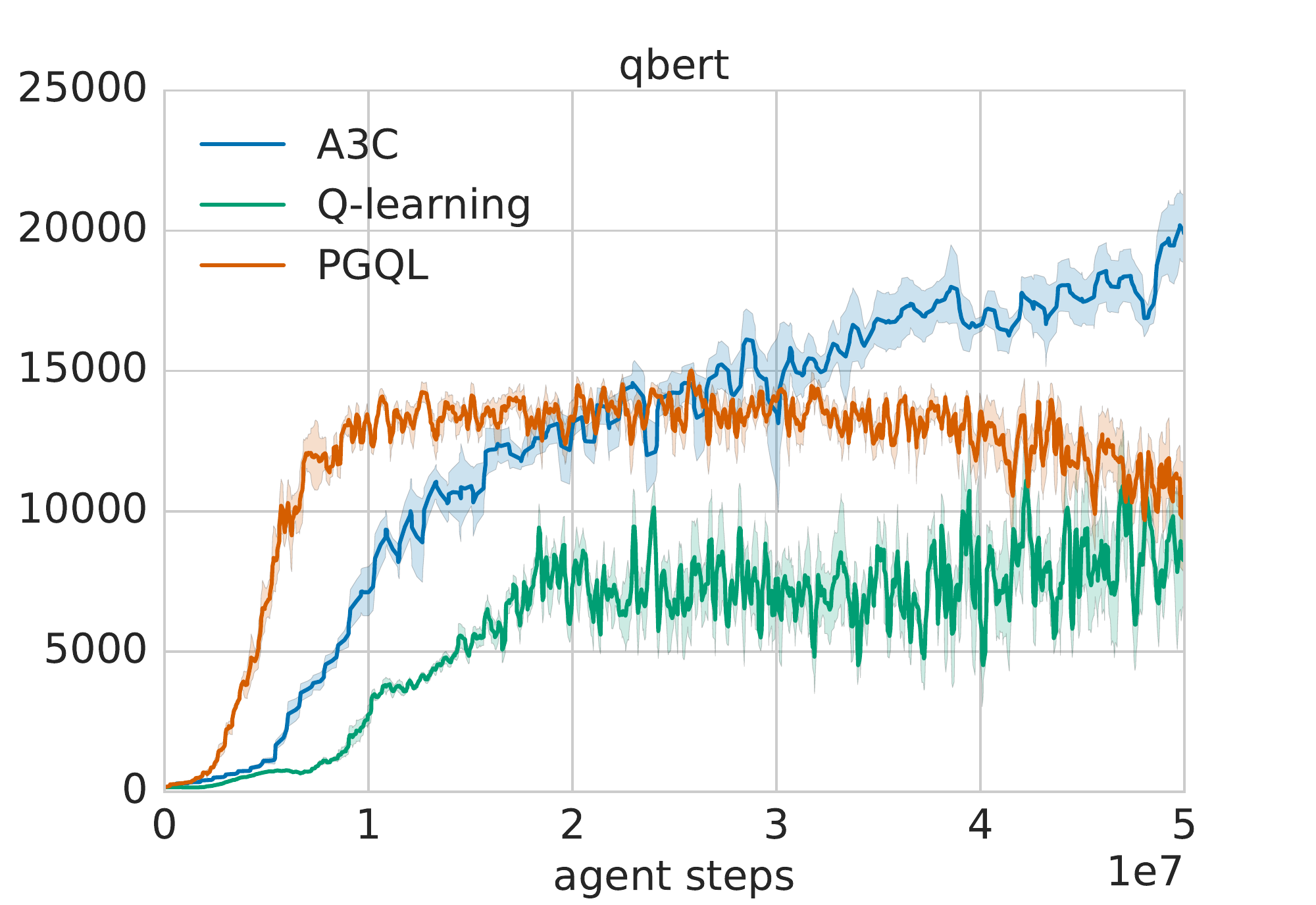}
    \end{subfigure}
    \begin{subfigure}[b]{0.4\textwidth}
        \includegraphics[width=\textwidth]{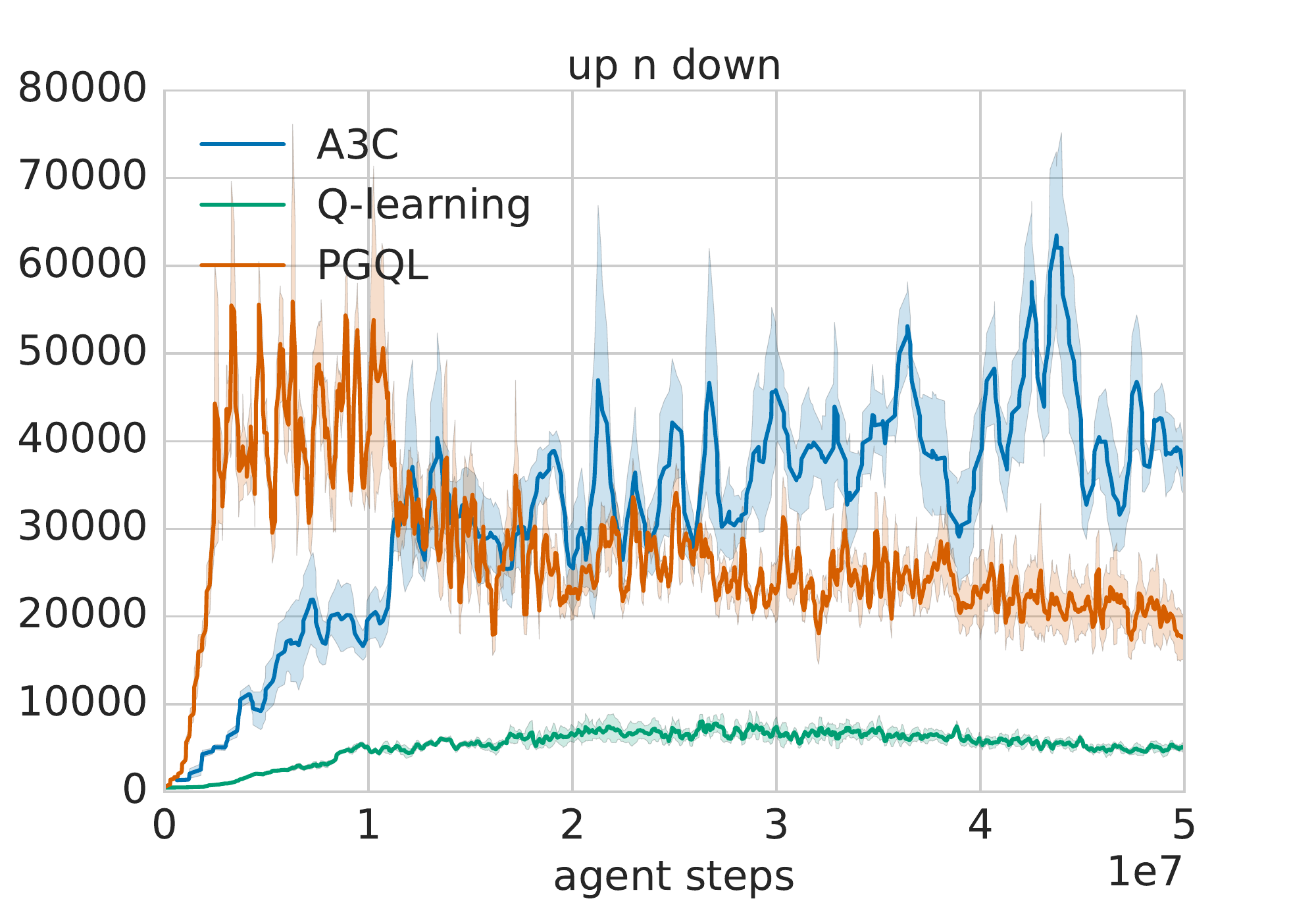}
    \end{subfigure}
  }
  \caption{Some Atari runs where PGQL performed poorly.
  \label{f-atari-failures}
  }
\end{figure}

It is worth noting that PGQL was the worst performer in only one game, in
cases where it was not the outright winner it was generally somewhere in between
the performance of the other two algorithms. Figure \ref{f-atari-successes}
shows some sample traces of games where PGQL was the best performer. In these
cases PGQL has far better data efficiency than the other methods.  In figure
\ref{f-atari-failures} we show some of the games where PGQL under-performed.
In practically every case where PGQL did not perform well it had better
data efficiency early on in the learning, but performance saturated or
collapsed. We hypothesize that in these cases the policy has reached a local
optimum, or over-fit to the early data, and might perform better were the
hyper-parameters to be tuned.

\begin{table}[h]
\begin{center}
\begin{tabular}{ c|c c c }
 & A3C & Q-learning & PGQL \\
\hline
\bf{Mean} & 636.8 & 756.3 & 877.2 \\
\bf{Median} & 107.3 & 58.9 & 145.6 \\
\end{tabular}
\end{center}
\caption{Mean and median normalized scores for the Atari suite from random
  starts, as a percentage of human normalized score.}
\label{t-atari-means-random}
\end{table}

\begin{table}[h]
\begin{center}
\begin{tabular}{ c|c c c }
  & A3C & Q-learning & PGQL \\
\hline
\bf{Mean} & 266.6 & 246.6 & 416.7 \\
\bf{Median} & 58.3 & 30.5 & 103.3 \\
\end{tabular}
\end{center}
\caption{Mean and median normalized scores for the Atari suite from human
starts, as a percentage of human normalized score.}
\label{t-atari-means-human}
\end{table}

\section{Conclusions}
We have made a connection between the fixed point of regularized policy gradient
techniques and the Q-values of the resulting policy. For small regularization
(the usual case) we have shown that the Bellman residual of the induced Q-values
must be small. This leads us to consider adding an auxiliary update to the
policy gradient which is related to the Bellman residual evaluated on a
transformation of the policy.  This update can be performed off-policy, using
stored experience.  We call the resulting method `PGQL', for policy
gradient and Q-learning.  Empirically, we observe better
data efficiency and stability of PGQL when compared to actor-critic or
Q-learning alone.  We verified the performance of PGQL on a suite of
Atari games, where we parameterize the policy using a neural network, and
achieved performance exceeding that of both A3C and Q-learning.


\section{Acknowledgments}
We thank Joseph Modayil for many comments and suggestions on the paper, and
Hubert Soyer for help with performance evaluation.  We would also like to thank
the anonymous reviewers for their constructive feedback.

\clearpage
\bibliographystyle{iclr2017_conference}
\bibliography{pgq}

\appendix

\section{PGQL Bellman residual}
Here we demonstrate that in the tabular case the Bellman residual of the induced
Q-values for the PGQL updates of (\ref{e-hybrid}) converges to zero as the
temperature $\alpha$ decreases, which is the same guarantee as vanilla
regularized policy gradient (\ref{e-rel-pol-grad}).
We will use the notation that $\pi_\alpha$ is the
policy at the fixed point of PGQL updates (\ref{e-hybrid}) for some
$\alpha$, \ie, $\pi_\alpha \propto \exp(\tilde Q^{\pi_\alpha})$, with induced
Q-value function $Q^{\pi_\alpha}$.

First, note that we can
apply the same argument as in \S \ref{s-bell-err} to show that $\lim_{\alpha
\rightarrow 0} \|\T^\star \tilde Q^{\pi_\alpha} - \T^{\pi_\alpha} \tilde
Q^{\pi_\alpha}\| = 0$ (the only
difference is that we lack the property that $\tilde Q^{\pi_\alpha}$ is the
fixed point of $\T^{\pi_\alpha}$). Secondly, from equation
(\ref{e-mod-fixed-pt}) we can write $\tilde Q^{\pi_\alpha} - Q^{\pi_\alpha}
= \eta (\T^\star \tilde Q^{\pi_\alpha} - Q^{\pi_\alpha})$.
Combining these two facts we have
\[
    \begin{array}{rcl}
      \|\tilde Q^{\pi_\alpha} - Q^{\pi_\alpha}\| &=& \eta \|\T^\star \tilde
      Q^{\pi_\alpha} - Q^{\pi_\alpha}\| \\
      &=& \eta \|\T^\star \tilde Q^{\pi_\alpha} -\T^{\pi_\alpha} \tilde Q^{\pi_\alpha} +
      \T^{\pi_\alpha} \tilde Q^{\pi_\alpha} - Q^{\pi_\alpha} \| \\
      &\leq& \eta ( \|\T^\star \tilde Q^{\pi_\alpha} -\T^{\pi_\alpha} \tilde Q^{\pi_\alpha}\| +
      \|\T^{\pi_\alpha} \tilde Q^{\pi_\alpha} - \T^{\pi_\alpha} Q^{\pi_\alpha} \| ) \\
      &\leq& \eta ( \|\T^\star \tilde Q^{\pi_\alpha} -\T^{\pi_\alpha} \tilde Q^{\pi_\alpha}\| +
      \gamma \|\tilde Q^{\pi_\alpha} - Q^{\pi_\alpha} \| ) \\
      &\leq& \eta / (1- \eta \gamma) \|\T^\star \tilde Q^{\pi_\alpha} -\T^{\pi_\alpha} \tilde
      Q^{\pi_\alpha}\|,
  \end{array}
\]
and so $\|\tilde Q^{\pi_\alpha} - Q^{\pi_\alpha}\| \rightarrow 0$ as $\alpha
\rightarrow 0$. Using this fact we have
\[
    \begin{array}{rcl}
      \|\T^\star \tilde Q^{\pi_\alpha} - \tilde Q^{\pi_\alpha}\| &=& \|\T^\star \tilde Q^{\pi_\alpha} -
      \T^{\pi_\alpha} \tilde Q^{\pi_\alpha} + \T^{\pi_\alpha} \tilde Q^{\pi_\alpha} - Q^{\pi_\alpha} + Q^{\pi_\alpha} - \tilde Q^{\pi_\alpha} \| \\
      &\leq&  \|\T^\star \tilde Q^{\pi_\alpha} -
      \T^{\pi_\alpha} \tilde Q^{\pi_\alpha}\| + \|\T^{\pi_\alpha} \tilde Q^{\pi_\alpha} - \T^{\pi_\alpha} Q^{\pi_\alpha} \| + \|Q^{\pi_\alpha} -
      \tilde Q^{\pi_\alpha} \| \\
      &\leq& \|\T^\star \tilde Q^{\pi_\alpha} -
      \T^{\pi_\alpha} \tilde Q^{\pi_\alpha}\| + (1+\gamma) \|\tilde Q^{\pi_\alpha} - Q^{\pi_\alpha} \| \\
      &<& 3 / (1 - \eta \gamma)\|\T^\star \tilde Q^{\pi_\alpha} -
      \T^{\pi_\alpha} \tilde Q^{\pi_\alpha}\|,
  \end{array}
\]
which therefore also converges to zero in the limit. Finally we obtain
\[
    \begin{array}{rcl}
      \|\T^\star Q^{\pi_\alpha} - Q^{\pi_\alpha}\| &=& \|\T^\star Q^{\pi_\alpha} - \T^\star \tilde Q^{\pi_\alpha} +
      \T^\star \tilde Q^{\pi_\alpha} - \tilde Q^{\pi_\alpha} + \tilde Q^{\pi_\alpha} - Q^{\pi_\alpha} \| \\
      &\leq& \|\T^\star Q^{\pi_\alpha} - \T^\star \tilde Q^{\pi_\alpha}\| +
      \|\T^\star \tilde Q^{\pi_\alpha} - \tilde Q^{\pi_\alpha}\| + \|\tilde Q^{\pi_\alpha} - Q^{\pi_\alpha} \| \\
      &\leq& (1 + \gamma) \| \tilde Q^{\pi_\alpha} - Q^{\pi_\alpha}\| +
      \|\T^\star \tilde Q^{\pi_\alpha} - \tilde Q^{\pi_\alpha} \|,
  \end{array}
\]
which combined with the two previous results implies that $\lim_{\alpha
\rightarrow 0}\|\T^\star Q^{\pi_\alpha} - Q^{\pi_\alpha} \| = 0$, as before.

\clearpage
\section{Atari scores}
\label{s-atari}
\begin{table}[h]
\begin{center}
\scriptsize
\begin{tabular}{ c|c|c|c }
Game  & A3C & Q-learning & PGQL\\
\hline
alien &      38.43 &      25.53 & \bf{46.70} \\
amidar &      68.69 &      12.29 & \bf{71.00} \\
assault &     854.64 &    1695.21 & \bf{2802.87} \\
asterix &     191.69 &      98.53 & \bf{3790.08} \\
asteroids &      24.37 &       5.32 & \bf{50.23} \\
atlantis &   15496.01 &   13635.88 & \bf{16217.49} \\
bank heist &     210.28 &      91.80 & \bf{212.15} \\
battle zone &      21.63 &       2.89 & \bf{52.00} \\
beam rider &      59.55 &      79.94 & \bf{155.71} \\
berzerk &      79.38 &      55.55 & \bf{92.85} \\
bowling &       2.70 &      -7.09 & \bf{3.85} \\
boxing &     510.30 &     299.49 & \bf{902.77} \\
breakout &    2341.13 & \bf{3291.22} &    2959.16 \\
centipede &      50.22 & \bf{105.98} &      73.88 \\
chopper command &      61.13 &      19.18 & \bf{162.93} \\
crazy climber & \bf{510.25} &     189.01 &     476.11 \\
defender &     475.93 &      58.94 & \bf{911.13} \\
demon attack & \bf{4027.57} &    3449.27 &    3994.49 \\
double dunk &    1250.00 &      91.35 & \bf{1375.00} \\
enduro & \bf{9.94} & \bf{9.94} & \bf{9.94} \\
fishing derby &     140.84 &     -14.48 & \bf{145.57} \\
freeway &      -0.26 & \bf{-0.13} & \bf{-0.13} \\
frostbite &       5.85 & \bf{10.71} &       5.71 \\
gopher &     429.76 & \bf{9131.97} &    2060.41 \\
gravitar &       0.71 &       1.35 & \bf{1.74} \\
hero & \bf{145.71} &      15.47 &      92.88 \\
ice hockey &      62.25 &      21.57 & \bf{76.96} \\
jamesbond &     133.90 &     110.97 & \bf{142.08} \\
kangaroo &      -0.94 &      -0.94 & \bf{-0.75} \\
krull &     736.30 & \bf{3586.30} &     557.44 \\
kung fu master &     182.34 & \bf{260.14} &     254.42 \\
montezuma revenge &      -0.49 & \bf{1.80} &      -0.48 \\
ms pacman &      17.91 &      10.71 & \bf{25.76} \\
name this game &     102.01 &     113.89 & \bf{188.90} \\
phoenix &     447.05 &     812.99 & \bf{1507.07} \\
pitfall &       5.48 & \bf{5.49} & \bf{5.49} \\
pong & \bf{116.37} &      24.96 & \bf{116.37} \\
private eye &      -0.88 & \bf{0.03} &      -0.04 \\
qbert & \bf{186.91} &     159.71 &     136.17 \\
riverraid &     107.25 &      65.01 & \bf{128.63} \\
road runner & \bf{603.11} &     179.69 &     519.51 \\
robotank &      15.71 & \bf{134.87} &      71.50 \\
seaquest &       3.81 &       3.71 & \bf{5.88} \\
skiing & \bf{54.27} &      54.10 &      54.16 \\
solaris &      27.05 & \bf{34.61} &      28.66 \\
space invaders &     188.65 &     146.39 & \bf{608.44} \\
star gunner &     756.60 &     205.70 & \bf{977.99} \\
surround &      28.29 &      -1.51 & \bf{78.15} \\
tennis & \bf{145.58} &     -15.35 & \bf{145.58} \\
time pilot &     270.74 &      91.59 & \bf{438.50} \\
tutankham &     224.76 &     110.11 & \bf{239.58} \\
up n down & \bf{1637.01} &     148.10 &    1484.43 \\
venture & \bf{-1.76} & \bf{-1.76} & \bf{-1.76} \\
video pinball &    3007.37 &    4325.02 & \bf{4743.68} \\
wizard of wor &     150.52 &      88.07 & \bf{325.39} \\
yars revenge &      81.54 &      23.39 & \bf{252.83} \\
zaxxon &       4.01 &      44.11 & \bf{224.89} \\
\hline
\end{tabular}
\end{center}
\caption{Normalized scores for the Atari suite from random starts, as a percentage
of human normalized score.}
\label{t-atari}
\end{table}

\end{document}